%% file: main.tex
\documentclass[10pt,twocolumn,letterpaper]{article}

 \usepackage{cvpr}              

\input{preamble}

\definecolor{cvprblue}{rgb}{0.21,0.49,0.74}
\usepackage[pagebackref,breaklinks,colorlinks,citecolor=cvprblue]{hyperref}
\usepackage{graphicx}
\usepackage{amssymb}
\usepackage{amsmath}
\usepackage{algorithm}
\usepackage{algpseudocode}
\usepackage{multirow}
\usepackage{array}
\usepackage{subcaption}
\usepackage{multicol}
\usepackage{wrapfig}

\newcommand{\down}[1]{\scriptsize\textcolor{red}{#1}}
\def\paperID{335} 
\def\confName{CVPR}
\def\confYear{2024}

\title{Ungeneralizable Examples}

\author{\bf Jingwen Ye \quad Xinchao Wang$^{\dagger}$ \thanks{ $^{\dagger}$ Corresponding author.}\\
 National University of Singapore\\
{\tt\small jingweny@nus.edu.sg, xinchao@nus.edu.sg}
}
\makeatletter
\def\thanks#1{\protected@xdef\@thanks{\@thanks
\protect\footnotetext{#1}}}
\makeatother
\begin{document}
\maketitle
\input{sec/0_abstract}    
\input{sec/1_intro}
\input{sec/2_related}
\input{sec/3_method}
\input{sec/4_experiments}
\input{sec/5_conclusion}
\input{sec/acknowledge}
{\small
\bibliographystyle{ieeenat_fullname}

\input{main.bbl}
}

\input{sec/X_suppl}

\end{document}

%% file: preamble.tex
\usepackage[dvipsnames]{xcolor}


%% file: sec/0_abstract.tex
\begin{abstract}

The training of contemporary deep learning models heavily relies on publicly available data, posing a risk of unauthorized access to online data and raising concerns about data privacy. Current approaches to creating unlearnable data involve incorporating small, specially designed noises, but these methods strictly limit data usability, overlooking its potential usage in authorized scenarios.
In this paper, we extend the concept of unlearnable data to conditional data learnability and introduce \textbf{U}n\textbf{G}eneralizable \textbf{E}xamples (UGEs). UGEs exhibit learnability for authorized users while maintaining unlearnability for potential hackers. The protector defines the authorized network and optimizes UGEs to match the gradients of the original data and its ungeneralizable version, ensuring learnability. To prevent unauthorized learning, UGEs are trained by maximizing a designated distance loss in a common feature space. Additionally, to further safeguard the authorized side from potential attacks, we introduce additional undistillation optimization.
Experimental results on multiple datasets and various networks demonstrate that the proposed UGEs framework preserves data usability while reducing training performance on hacker networks, even under different types of attacks.
\end{abstract}

%% file: sec/1_intro.tex
\section{Introduction}
\label{sec:intro}

The widespread availability of `free' internet data has played a pivotal role in advancing deep learning and computer vision models. However, a notable concern arises from the collection of datasets without explicit consent, with personal data often gathered unknowingly from the internet. This practice has raised public concerns about the potential unauthorized and, in some cases, potentially illegal exploitation of personal information. These issues have gained even greater significance with the introduction of the General Data Protection Regulation (GDPR) by the European Union, placing a renewed emphasis on data protection within the AI community.

\begin{figure}[t]
\centering
\includegraphics[scale = 0.55]{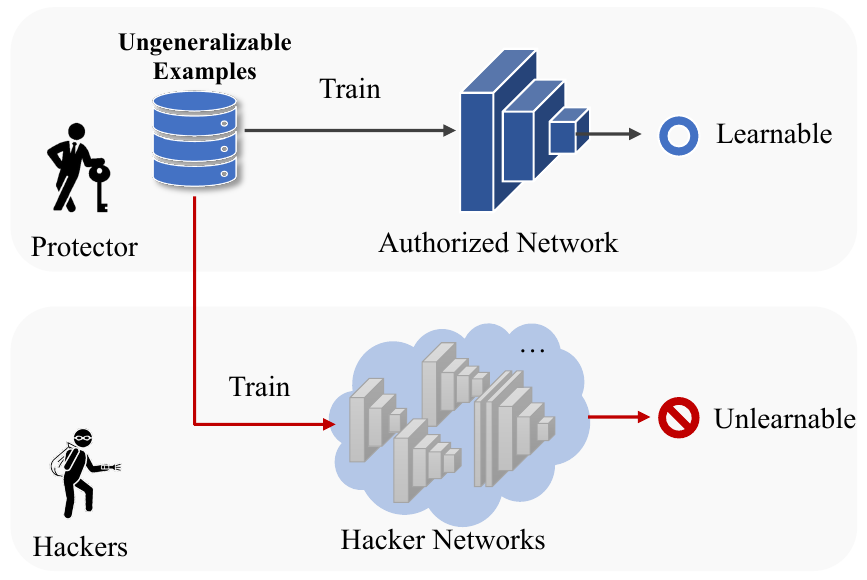}
\caption{
The threat model of ungeneralizable examples involves generating UnGeneralizable Examples. Once created, both the protector and the hacker gain access to the UGEs rather than the original data. While the UGEs can effectively train the protector's network, they result in a performance drop on hacker networks.
}
\label{fig:main}
\end{figure}

To address the risk of machine learning models capturing private data, recent developments have focused on the concept of unlearnable examples (ULE)~\cite{huang2020unlearnable,fu2021robust,ren2022transferable}. Unlearnable examples represent data types that deep learning models struggle to effectively learn useful information from. A common method for generating unlearnable examples involves a min-min bilevel optimization framework, deceiving the model into learning a false connection between noise and labels. Consequently, models trained on such unlearnable examples exhibit significantly reduced performance, emphasizing the importance of robust data protection in machine learning.
It's crucial to note that, oftentimes, the data itself is not inherently problematic; instead, \textit{it is the manner in which they are utilized that demands careful consideration}. Therefore, we argue that such an across-the-board data protection rule could address some stringent privacy issues but might impede the shareable community under normal service conditions.

In our study, we broaden the conventional assumption in existing ULE methods, introducing a more adaptable and pragmatic data protection paradigm referred to as ungeneralizable examples. In contrast to the conventional ULE framework, we posit that the data can be learnable by networks pre-defined by the protector. This approach enables the protector to maintain authorized usage of the collected data, addressing the inflexible concern of unlearnable examples. Moreover, it offers an alternative for the protector when they need to share their data for specific legitimate purposes. The fundamental concept of UGEs is visually represented in Figure~\ref{fig:main}.

In UGE, the protector pre-defines the authorized network before generating the ungeneralizable version of the data. We approximate the training trajectories of the original data and the ungeneralizable data to ensure that the data's learnability remains unchanged. To prevent the data from being learned by hackers, we maximize the feature distance in the common feature space, where unlearnability can transfer to multiple hacker networks. Additionally, to further enhance the confidentiality of the ungeneralizable examples, we introduce the undistill loss, aiming to prevent hackers from recovering the original data from the protector network.

In summary, the contributions of this paper can be outlined as follows:
\begin{itemize}
    \item \textit{\textbf{Introduction of Ungeneralizable Examples Paradigm}}: We propose a versatile data protection paradigm termed ungeneralizable examples. This paradigm enables the legitimate use of data by the protector while preventing unauthorized usage by potential hackers. It introduces a pragmatic scenario, challenging the unauthorized training of machine learning models.
    \item \textit{\textbf{Innovative Solution for Learnability and Unlearnability Switchover}}: We introduce a novel approach to address the switch between data learnability and unlearnability using three distinct losses. This is the first and only method, to our knowledge, that achieves this switchover effectively.
    \item \textit{\textbf{Empirical Verification of Effectiveness and Robustness}}: We empirically verify the effectiveness of our proposed approach with different network backbones on diverse datasets. Furthermore, we assess its robustness under various network architectures and multiple types of attacks.
\end{itemize}

%% file: sec/2_related.tex
\section{Related Work}
\label{sec:related}
\subsection{Data Privacy Protection}
Ensuring data privacy protection is crucial for safeguarding individuals' sensitive information, preserving autonomy, and fostering trust in the digital landscape. This commitment is instrumental in the ethical and responsible development of technology. 
Here we group the data privacy protection into visual information protection and data protection from machine learning.

The essence of visual information protection lies in rendering data visually unrecognizable or inaccessible to third parties.
A direct approach involves employing basic techniques like pixelization, blurring, or scrambling to obscure facial features in images. Alternatively, recent advancements explore the use of encryption~\cite{kaur2020comprehensive,wang2019fast} directly applied to an image, followed by inpainting~\cite{xiang2023deep,lugmayr2022repaint,wang2023imagen}, making it challenging to recover the original content.
As another illustration, the concept of dataset condensation~\cite{cazenavette2022distillation,liu2023slimmable,SonghuaNeurIPS22,SonghuaNeurIPS23,RuoNanTPAMI24} is introduced to distill the essence of data into a compact synset. This approach aims to safeguard the original data's integrity while retaining its ability to effectively train a neural network.
Regarding data privacy, federated learning~\cite{zhang2021survey,li2021model,tan2022towards} emphasizes a distributed model training paradigm that prioritizes keeping sensitive information localized on individual devices, thereby mitigating privacy risks associated with centralized data storage or sharing.

Data protection from machine learning primarily focuses on the control and management of learnable features extracted from networks.
For exmaple, machine unlearning~\cite{bourtoule2021machine,tarun2023fast,liu2023muter} exemplifies the recalibration of machine learning models through the selective discarding of specific data points, patterns, or predictions. This process involves the removal of sensitive data information from the network, effectively eliminating the risk of unintended data exposure.
An additional aspect of data protection involves preventing data from being learned by machine unlearning models. Huang et al.~\cite{huang2020unlearnable} have made a significant contribution to safeguarding image data from unauthorized machine learning exploitation. They introduced a method focused on generating error-minimizing noise with the primary goal of intentionally degrading images uploaded to the internet. This degradation aims to impede the training process of neural networks. As a result, images incorporating this introduced noise are classified as unlearnable examples~\cite{fu2021robust,ren2022transferable}.

We posit our proposed ungeneralizable examples as an expanded version of unlearnable examples, offering enhanced flexibility in data management. In this approach, the data remains unlearnable by the defender while remaining learnable by the protector, thereby providing a more nuanced control over the learning dynamics.

\subsection{Model Privacy Protection}
In the realm of model privacy protection, our focus centers on Intellectual Property (IP) safeguarding. The escalating commercial significance of deep networks has garnered heightened attention from both academia and industry, emphasizing the imperative for robust IP protection.

As a conventional technique, network watermarking~\cite{li2022encryption,szyller2021dawn,le2020adversarial} involves embedding identification information into the target network, enabling copyright claims without compromising the network's predictive capabilities.
Numerous recent studies~\cite{kariyappa2020defending,li2022defending,Ye2023PartialNC} have investigated defensive strategies against model stealing, aiming to safeguard the intellectual property of the network.
As an additional measure for intellectual property (IP) protection, knowledge undistillation~\cite{ma2020undistillable,ye2022safe} is introduced to prevent knowledge theft by other networks. This entails maintaining the network's fundamental prediction performance while inducing a performance drop when attempting to distill knowledge.

Our proposed ungeneralizable examples have something common with knowledge undistillaion, which are designed to introduce modifications to the original images, leading to suboptimal performance on unauthorized networks while preserving their efficacy in training the protector's network.

\subsection{Adversarial and Data Poisoning Attacks}
Adversarial attacks~\cite{Madry2017TowardsDL,Dong2017BoostingAA,guo2019simple,Akhtar2018ThreatOA,Zhang2021ASO, ye2024mutual} are designed to deceive machine learning models by adding small, imperceptible perturbations to input data, causing the model to generate incorrect outputs or misclassify inputs. One of the traditional attack methods~\cite{Goodfellow2014ExplainingAH} is to use gradient information to update the adversarial example in a single step along the direction of maximum classification loss.

Data poisoning~\cite{zhang2020online,schwarzschild2021just,tolpegin2020data} is a type of adversarial attack that involves manipulating the training data used to train machine learning models. The goal of these attacks is to introduce malicious or misleading data into the training set, with the intention of influencing the performance of the trained model.

However, such methods don't affect the model's performance on clean data, which makes them unsuitable for data privacy pretection.

%% file: sec/3_method.tex
\section{Proposed Method}
\textbf{Assumptions on Protector’s Capability}:
We assume that the protector has unrestricted access to the specific dataset they intend to make ungeneralizable. However, it's crucial to clarify that the protector lacks the capacity to interfere with the training process and does not have access to the entire training dataset. In simpler terms, the protector's influence is confined to transforming their designated data portion into ungeneralizable examples.
Furthermore, it's essential to underscore that once the ungeneralizable examples are generated, the protector is prohibited from making further modifications to their data. Importantly, these modifications are irreversible. In other words, once the alterations are applied, the original data is replaced by the modified versions.

\begin{figure*}[t]
\centering
\includegraphics[width = 1\textwidth]{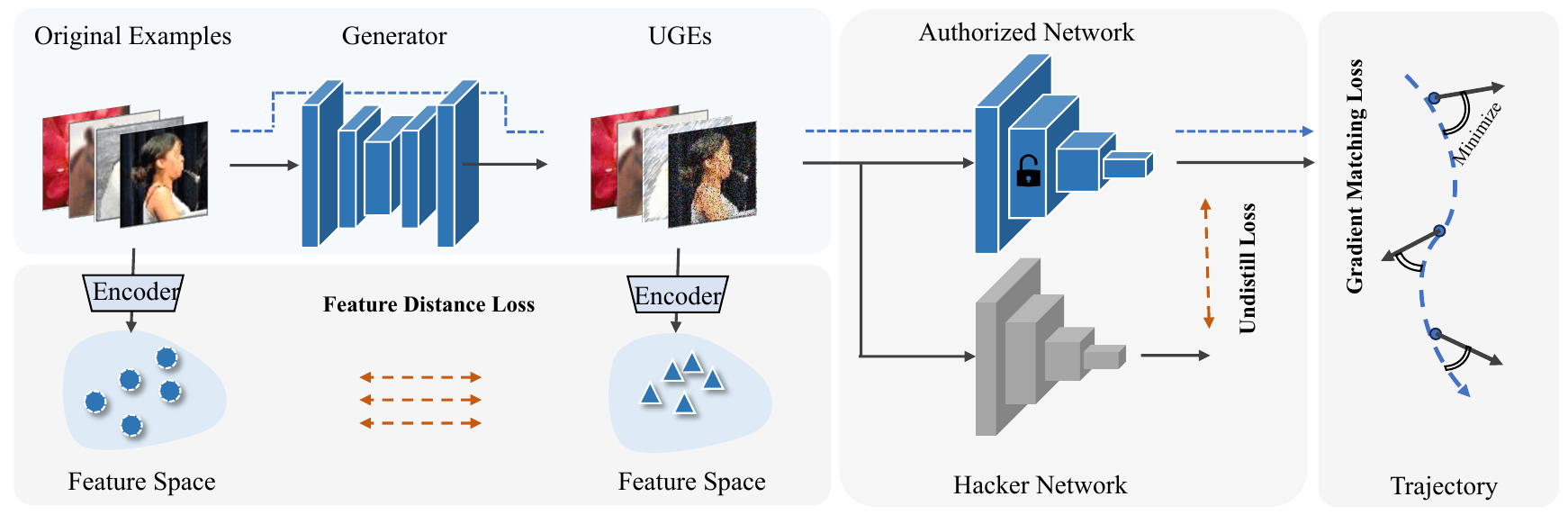}
\caption{
The comprehensive workflow of UGEs involves the protector training a generator to produce the ungeneralizable version of the original examples. Three distinct loss functions are employed in training the generator: gradient matching loss, feature distance loss, and undistill loss. Upon completion of the training process, the UGEs are published, and both the protector and hackers no longer have access to the original examples.
}
\label{fig:frame}
\end{figure*}

\subsection{Problem Formulation}
Following the previous setting on unlearnable examples~\cite{huang2020unlearnable},
we focus on image classification tasks in this paper.

Suppose $\mathcal{D}=\{(x,y)\}^{(n)}$ is a clean  training dataset with $K$-class,
where images can be denoted as $x\in \mathcal{X} \subset \mathbb{R}^d$, the corresponding groundtruth labels are denoted as $y\in \mathcal{Y} =\{1,2,...,K\}$. 
Two distinct networks are introduced: the authorized network, denoted as $f_\theta$, and the hacker's network, denoted as $f'_{\theta_A}$. The network $f_\theta$ is predetermined by the protector, where the network's architecture and initial parameters are set. 
Alongside the original data $\mathcal{D}$, the protector utilizes $f_\theta$ to generate the ungeneralizable version of the dataset, denoted as $\mathcal{D}_u=\{(x_u,y_u)\}^{(n)}$. This process is defined as follows:
\begin{equation}
    x_u\leftarrow x+\delta(f_\theta),\quad  y_u \leftarrow y; \quad \{(x,y)\} \in \mathcal{D}.
\end{equation}
Here $\delta(f_\theta) \subset \mathbb{R}^d$ is the generated ungeneralizable noise that is related to the authorized network $f_\theta$. The ungeneralizable noise is typically regulated to be imperceptible.
We omit $f_\theta$ from the ungeneralizable noise in the rest of the paper.  
The ungeneralizable dataset $\mathcal{D}_u$ is assumed to be the shareable dataset collected by both the hackers and the protector, which will be utilized to train both the authorized network $f_\theta$ and the hacker network $f'_{\theta_A}$.

The generation of ungeneralizable examples serves two main objectives: firstly, they are designed to remain learnable for the authorized network $f_\theta$; secondly, they are intended to become unlearnable for the malicious networks $f'_{\theta_A}$ employed by the hackers. Thus, the objective could be formulated as:
\begin{equation}
\begin{split}
        \underset{\theta}{\min}\frac{1}{n}\sum_{(x,y)\in \mathcal{D}}^n \underset{\|\delta\|\le \rho}{\min}&\Big[\mathcal{L}\big(f'_{\theta_A}(x+\delta),y\big) + \\
        \big\|\mathcal{L}\big(f_\theta(x&+\delta),y\big)-\mathcal{L}\big(f_\theta(x),y\big)\big\| \Big],
\end{split}
\label{eq:problem}
\end{equation}
where $\mathcal{L}(\cdot)$ denotes the loss function for network training, and $\rho$ is the radius represents the radius of the applied ungeneralizable noise.
The formal section of the objective function introduces error-minimizing noise to render the data unlearnable by diminishing the associated training loss, making it challenging for hackers using $f'_{\theta_A}$ to acquire the knowledge. The latter part of the objective function aims to reconstruct this knowledge on the authorized network $f_\theta$ by minimizing the training loss between the clean input and the ungeneralizable version input.

\textbf{Design Goals of UGEs}:
We aim to generate the ungeneralizable examples with the following characteristics:
\begin{itemize}
\item \textbf{Visual Integrity}: The ungeneralizable version of the images should remain visually recognizable to human observers, meaning that the ungeneralizable noise should be confined to a small norm.
\item \textbf{Effectiveness}. UGEs facilitate authorized training on the authorized networks while preventing unauthorized training by hackers, demonstrating conditional learnability.
\item \textbf{Robustness}.  The unlearnability of UGEs should be stable and resistant to attacks by hackers; their safety should be verified under various types of attacks. Additionally, it should be transferable to different network architectures.
\item \textbf{User-friendliness}.  It should be convenient for authorized usage. That is, it shouldn't affect the training process on the authorized network. No new losses or components are introduced for training on UGEs, and it shouldn't increase the computational load of the training process.
\end{itemize}
\subsection{Ungeneralizable Examples}
As is shown in Fig.~\ref{fig:frame}, the framework of obtaining the ungeneralizable version of the original data is depicted, where we train a generator to synthesize the UGEs:
\begin{equation}
   x_u \leftarrow Clamp\big(\mathcal{G}(x), x-\rho, x+\rho \big)\quad x\in \mathcal{D},
\end{equation}
where the $Clamp()$ operation is to constrain the ungeneralizable noise's norm within $\rho$. 
A total of three loss functions are utilized to train the generator $\mathcal{G}$:
\begin{equation}
    \mathcal{L}_{all}= \mathcal{L}_{gm} + \lambda_{fd}\cdot\mathcal{L}_{fd} + \lambda_{ud}\cdot\mathcal{L}_{ud},
\end{equation}
where $\mathcal{L}_{gm}$ is the gradient matching loss to ensure UGEs learnanle on the authorized network, $\mathcal{L}_{fd}$ is the feature distance loss to make UGEs unlearnable on the hacker networks, and $\mathcal{L}_{ud}$ is the undistill loss to make the original examples inreversible on the authorized network. $\lambda_{fd}$ and $\lambda_{ud}$ are the weights to balance each loss item.

\textbf{Learnable on the authorized network.}
As is stated in Eq.~\ref{eq:problem}, the latter loss item which tries to minimize the training loss between the inputs of $x$ and $x_u$.
Note the architecture and the initial parameters $\theta_0$ of the authorized network are confirmed, 
the training process of $f_\theta$ on the original dataset $\mathcal{D}$ could be determined:
\begin{equation}
f: \theta_{t+1}\leftarrow \theta_{t}-\eta\frac{1}{n} \sum_{(x,y)\in\mathcal{D}} \nabla\mathcal{L}\big(f_{\theta_t}(x),y\big),
\end{equation}
where $\eta$ is the learning rate, and $t = \{0,1,...,T-1\}$ is the training epoch number, $T$ is the total training epoch number. $\nabla \mathcal{L}$ is the gradients while in each training epoch.

To sustain the learning trajectory on the original data $x$, we introduce the gradient matching loss during the training process between $x$ and $x_u$. Specifically, we randomly sample several intermediate training epochs from $\{0,1,...,T-1\}$, denoted as $\tau$. The gradient matching loss is then calculated in these sampled epochs, thus the gradient matching loss can be finally expressed as:
\begin{equation}
\begin{split}
  \mathcal{L}_{gm} =\frac{1}{|\tau|\!\times\! n}\sum_{t\in \tau} \sum_{(x,y)\in \mathcal{D}}\mathcal{D}{ist}&\big[\nabla\mathcal{L}\big(f_{\theta_t}(x),y\big),\\
  &\nabla\mathcal{L}\big(f_{\theta_t}(x_u),y\big)\big],   
\end{split}
\label{eq:gm}
\end{equation}
where $\mathcal{D}{ist}(\cdot)$ represents the cosine distance, and we employ it to distill the gradient information from $x$ to $x_u$.
Minimizing the gradient matching loss $\mathcal{L}_{gm}$ effectively aligns the training trajectory of the original data with that of the ungeneralizable data. This optimization ensures the preservation of data learnability.

\textbf{Unlearnable on the hacker network.}
As outlined in Eq.~\ref{eq:problem}, the first loss renders the data unlearnable on hacker networks. Traditional unlearnable methods address this optimization challenge by introducing error-minimizing noise, typically through bi-level optimization, which is considered less efficient.

In this approach, we leverage a shared feature space to apply perturbations, thereby achieving the unlearnable characteristic in the data. As depicted in Fig.~\ref{fig:frame}, a common image encoder $\mathcal{E}_i$ extracts features from both the original data $\mathcal{D}$ and its ungeneralizable version $\mathcal{D}_u$. To ensure that the feature perturbation designed in this feature space remains effective across diverse networks, a robust and powerful encoder selection becomes crucial. 

Considering this perspective, we opt to utilize the pre-trained image encoder of the CLIP model~\cite{radford2021learning}. As a leading Vision-and-Language (VL) model, CLIP learns state-of-the-art image representations from scratch on a dataset containing 400 million image-text pairs collected from the internet. This training allows it to excel in various tasks, including zero-shot classification. Alongside the powerful image encoder $\mathcal{E}_i$ offered by CLIP, an additional textual encoder $\mathcal{E}_t$ is available to provide supplementary guidance.

To be specific, the feature distance loss $\mathcal{L}_{fd}$ can be computed as follows:
\begin{equation}
\begin{split}
\!\mathcal{L}_{fd}\! =\!\frac{1}{n}\!\sum_{x_u\in \mathcal{D}_u}\! \ell_{feat}(x_u,\mathcal{E}_i,\mathcal{D}) + \ell_{tri}(x_u,\mathcal{E}_i,\mathcal{E}_t,\mathcal{D}),\\
\end{split}
\label{eq:fd}
\end{equation}
which contains two main loss items $\ell_{feat}$ and $\ell_{tri}$. The former loss $\ell_{feat}$ pushes the features of the ungeneralizable $x_u$ away from the original $x$:
\begin{equation}
\begin{split}
    \ell_{feat}(x_u,\mathcal{E}_i,\mathcal{D}) &= -\|\mathcal{F}_i-\mathcal{F}_i'\|^2,\\
where \quad \mathcal{F}_i =\frac{\mathcal{E}_i(x)}{\|\mathcal{E}_i(x)\|}, & \mathcal{F}'_i=\frac{\mathcal{E}_i(x_u)}{\|\mathcal{E}_i(x_u)\|},
\end{split}
\end{equation}
where $\mathcal{F}_i$/$\mathcal{F}'_i$ is the normalized features and $\ell_{feat}$ is calculated based on the MSE loss.

In addition to maximizing the similarity between the features of the original input and the ungeneralizable input, we introduce an additional triplet loss. This triplet loss ensures that the features of ${\mathcal{F}'_i}$ in the ungeneralizable input can be effectively transferred to various hacker networks.
\begin{equation}
\scalebox{0.9}{$
\begin{split}
\!\ell_{tri} (x_u,\mathcal{E}_i,\mathcal{E}_t,\mathcal{D})\! =\!\|\mathcal{F}'_i-\mathcal{F}'_t&\|^2\!+\!\max\big(0, \alpha\!-\!\|\mathcal{F}_i'\!-\!\mathcal{F}_t\|^2\big),\\
where \quad \mathcal{F}_t = \frac{\mathcal{E}_t(y)}{\|\mathcal{E}_t(y)\|}, \mathcal{F}'_t &= \underset{\mathcal{F}_t^c}{\arg\min} Sim(\mathcal{F}_i,\mathcal{F}_t^c).
\end{split}
$}
\end{equation}
Here, $\alpha$ is the margin of the triplet loss and  $\mathcal{F}_t$ is the textual features with the groundtruth label $y$ as input and $Sim(\cdot)$ is the similarity function measuring the distance between the textual features and image features. $\mathcal{F}_t^c$ is the textual input with label $c$ ($c\in \{1,2,...,K\}$ and $c\ne y$). Consequently, $\mathcal{F}'_t$ refers to the textual features with the least similarity to the original image encoder features $\mathcal{F}_i$.
The term $\ell_{tri}$ encourages the features of ungeneralizable examples to move away from their associated textual features towards those of the least similar textual features.

\textbf{Untransferable on the authorized network.}
After the ungeneralizable version of the data $\mathcal{D}_u$ is published, both the protector and hackers gain access to UGEs. 
UGEs show to be unlearable on the hacker networks when standard training with $\min_{\theta_A} \mathcal{L}(f'_{\theta_A}(x_u),y_u)$.
On the contrary, UGEs can be employed for normal training on the network $f_\theta$ authorized by the protector. By minimizing the gradient matching loss $\mathcal{L}_{gm}$ as defined in Eq.~\ref{eq:gm}, $f_\theta$ attains similar performance to when trained with the original data $\mathcal{D}$.
The protector doesn't constrain the authorized network to be confidential, which means the hackers also have access to the authorized network $f_\theta$ (including architecture and parameters).
In this way, the hackers have another alternative to train their networks, with both the ungeneralizable examples $\mathcal{D}_u$ and $f_\theta$ available. 

To be concrete, the protector has authorized the data learning on the authorized network $f_{\theta}$,  and there exists a potential risk for hackers to exploit distillation-based learning process, expressed as:
\begin{equation}
f': \underset{\theta_A}{\min} \frac{1}{n}\sum_{x_u\in \mathcal{D}_u}\mathcal{L}_{kd} \big(f_\theta(x_u), f'_{\theta_A}(x_u)\big),
\label{eq:distill}
\end{equation}
where $\mathcal{L}_{kd}$ represents the KL-divergence loss for distilling knowledge directly from the authorized network $f_\theta$. This poses a significant security risk as it could expose the confidentiality of UGEs through the authorized network $f_\theta$.

Considering this concern, we introduce an undistill loss $\mathcal{L}_{ud}$ to safeguard the knowledge of the authorized network. Building upon prior work on knowledge undistillation~\cite{ma2020undistillable} designed for network IP protection, our proposed undistill loss is expressed as:
\begin{equation}
\scalebox{0.9}{$
\begin{split}
\! \underset{\mathcal{D}_u}{\min} \mathcal{L}_{ud}\! =\! \frac{1}{n}\!\sum_{(x_u,y_u)\in \mathcal{D}_u}\!\big[ &\mathcal{L}(f_\theta(x_u), y_u)
    \! -\!  \omega \mathcal{L}_{kd} \big(f_\theta(x_u), f'_{\theta_A}(x_u)\big)\big],
\end{split}
\label{eq:ud}
$}
\end{equation}
where $\mathcal{L}(\cdot)$ represents the standard training loss of $f_\theta$ and $\omega$ is the balancing weight. It's worth noting that in previous knowledge undistillation approaches, the undistill loss is employed to update the parameters of the network to be protected. In our case, we maintain $f_\theta$ and $f'_{\theta_A}$ fixed and optimize $x_u$ to ensure that its learnable knowledge within $f_\theta$ cannot be transferred to $f'_{\theta_A}$. This additional optimization step further enhances the security of our proposed ungeneralizable examples.

It's essential to highlight that in this context, {\textit{we do not restrict $f'_{\theta_A}$ to any specific networks}}; it can be any arbitrarily initialized network. Our proposed UGEs not only demonstrate effectiveness on the randomly chosen $f'_{\theta_A}$ but also exhibit generalizability, extending their unlearnability characteristics to other networks.

\subsection{Algorithm}
The whole algorithm is depicted in Alg~\ref{alg::attack}.

\begin{algorithm}[h]
\caption{The framework of the proposed UGEs.}
\small
\label{alg::attack}
\begin{algorithmic}[1]
\Require 
$\mathcal{D}$: original data to be protected;
$f_\theta$: authorized network;
$\{\theta_0,\theta_1,...,\theta_\tau\}$: sampled trajectory of the authorized network.
$f'_{\theta_A}$: randomly initialized hacker network.
$\rho$: ungeneralizable noise $\ell_{\infty}$ bound;
\Ensure $\mathcal{D}_u$: ungeneralizable examples.
\State Initialize the generator model $\mathcal{G}$;
\State Initialize the text input as `A photo of a $<\text{CLASS}>$';
\State Input text input to $\mathcal{E}_t$ to get textual features for all classes;
\While{\textnormal{not convergence}}
\State Input $x$ to $\mathcal{G}$ and get $x_u = \mathcal{G}(x)$ bounded with $\rho$;
\State Randomly choose $\theta_t$ from the input trajectory;
\State Input both $x$ and $x_u$ to $f_{\theta_t}$ to calculate $\mathcal{L}_{gm}$ with Eq.~\ref{eq:gm};
\State Input both $x$ and $x_u$ to encoder $\mathcal{E}_i$ to get $\mathcal{F}_i$ and $\mathcal{F}'_i$ ;
\State Calculate $\mathcal{L}_{fd}$ with Eq.~\ref{eq:fd};
\State Input $x_u$ to both networks $f_\theta$ and $f'_{\theta_A}$;
\State Calculate $\mathcal{L}_{ud}$ with Eq.~\ref{eq:ud};
\State Update $\mathcal{G}$ by minimizing  $\mathcal{L}_{all}$;
\EndWhile
\State Get the ungeneralizable examples $\mathcal{D}_u$ with $\mathcal{G}$;
\State Publish the final $\mathcal{D}_u$.
\end{algorithmic}
\end{algorithm}

\subsection{UGEs in Various Usages}
\label{sec:use}
The proposed UGEs seamlessly combine both data learnability and unlearnability within a single framework, showcasing a flexible approach to data management suitable for various applications.

\textit{\textbf{Scenario I}: Utilizing UGEs in Decentralized Model Training.}
In scenarios resembling federated learning, where privacy constraints exist in individual local servers, UGEs offer a viable solution. The global server establishes the initial global model, communicates the model information to each local server, and enables the joint training of the global model. UGEs effectively address privacy concerns by selectively publishing data for specific use cases.

\textit{\textbf{Scenario II}: Enhancing Code Publication Safety with UGEs.}
In open-source platforms such as GitHub, researchers are encouraged to share their code for collaborative AI development. However, instances arise where researchers collectively publish their gathered data. To mitigate the risk of malicious utilization of this data, researchers can opt to publish the ungeneralizable version of their training data, ensuring a more secure sharing environment.

\textit{\textbf{Scenario III}: Ensuring Secure Data Transmission with UGEs.}
In instances where secure data transmission is required to train a downstream network, a secure process can be established. The receiver initiates the transmission by sending its information to the protector. Subsequently, only the UGEs are transmitted to the receiver, mitigating the risk of interception by hackers during the transmission process. This approach ensures a secure and protected data exchange, with UGEs playing a pivotal role in safeguarding sensitive information.

%% file: sec/4_experiments.tex
\section{Experiments}
In this section, we conduct comprehensive experiments to validate the effectiveness of the robust ungeneralizable examples. 
Additional details regarding the experiment setup can be found in the supplementary materials.

\subsection{Experiment Setup}

\textbf{Datasets.}
Continuing the experimental setup from previous unlearnable methods, we present our results on CIFAR-10, CIFAR-100, and TinyImageNet datasets. The input size for CIFAR-10 and CIFAR-100 datasets is $32 \times 32$, while for the TinyImageNet dataset, we utilize an input size of $256 \times 256$.

\textbf{Model Training.}
We employ the PyTorch framework for implementation and investigate several network backbones, including plain CNN, LeNet, ResNet, MobileNetV2, and ShuffleNetV2. The generator utilizes a ResNet backbone. 

In our supposition, both the authorized network and hacker networks are optimized using standard Stochastic Gradient Descent (SGD). The authorized network $f_\theta$ is determined with given network architecture and initialization parameters.
We assume networks with either different architectures or different initialization parameters are regarded as hacker networks. When training the UGEs, we randomly select a distinct network as the hacker network, exclusively including the hacker network for training.

\textbf{Evaluation Metrics.}
We evaluate the data protection capability of the ungeneralizable noise using test accuracy. A low test accuracy on hacker networks indicates that the model has learned minimal knowledge from the training data, reflecting strong protection. Conversely, a high test accuracy on the authorized network indicates that the model has successfully learned knowledge from the training data, demonstrating data learnability for authorized usage.

\begin{table*}[t]
\small
\caption{Experimental Results on CIFAR-10, CIFAR-100 and TinyImageNet datasets , where ResNet-18 is used as the backbone of the authorized network. Acc changes are shown in red comparing with the network normal trains on the original dataset.}
\begin{subtable}{1\linewidth}
\centering
\begin{tabular}{ccccp{18mm}<{\centering}p{18mm}<{\centering}p{18mm}<{\centering}p{18mm}<{\centering}}
\toprule
\multirow{2}{*}{\textbf{Dataset}} & \multirow{2}{*}{\textbf{Method}} & \multirow{2}{*}{\textbf{ Scheme}} & \multirow{2}{*}{\textbf{Acc. (Authorized)}} & \multicolumn{4}{c}{\textbf{Acc. (Hacker)}}  \\ \cmidrule{5-8} 
                           &                         &                                  &                                  & CNN & ResNetC-20 & ResNetC-32 & ResNet-18 \\ \midrule\midrule
CIFAR-10  & Original & Normal & 95.05 &86.57& 92.28 &93.04 &95.05  \\
CIFAR-10  & Original & Distill & -& 88.06 \down{(+1.49)} &92.62 \down{(+0.34)}  & 93.24 \down{(+0.20)}&  95.41 \down{(+0.36)}       \\ \midrule
CIFAR-10  & Unlearn & Normal  & 22.59 \down{(-72.46)} & 18.36 \down{(-68.21)} & 20.37 \down{(-71.91)} & 22.39 \down{(-70.65)}&22.44 \down{(-72.61)} \\
CIFAR-10 & Unlearn   & Distill   & - & 17.39 \down{(-69.18)}& 21.40 \down{(-70.88)}  & 22.08 \down{(-70.96)} & 21.48 \down{(-73.57}\\ \midrule
CIFAR-10  & UnDistill & Normal & 94.52 \down{(-0.47)}& 85.87 \down{(-0.70)}  & 85.91 \down{(-6.37)} &   86.98 \down{(-6.06)}& 88.07 \down{(-6.98)}\\
CIFAR-10 & UnDistill & Distill & -& 73.38 \down{(-13.19)}& 78.65 \down{(-13.63)} & 80.76 \down{(-12.28)} &84.07 \down{(-10.98)}\\ \midrule
CIFAR-10 & UGEs w/o UD & Normal  & 94.34 \down{(-0.71)} & 46.97 \down{(-39.6)}  &  56.97 \down{(-35.31)} & 75.07 \down{(17.97)}  &  45.95 \down{(-49.10)}   \\
CIFAR-10 & UGEs w/o UD & Distill  &-  &  75.23 \down{(-11.34)}& 69.08 \down{(23.20)}    &  77.45 \down{(-15.59)}  & 87.10 \down{(-7.95)} \\ \midrule
CIFAR-10  & UGEs  & Normal  &  93.89 \down{(-1.16)}  &   26.46 \down{(-60.11)} &  30.63\down{(-61.65)}  &36.08 \down{(-56.96)}   & 26.12 \down{(-68.93)}   \\
CIFAR-10  & UGEs & Distill  & - & 32.08 \down{(-54.49)}   &  37.22\down{(-55.06)}   &  47.59 \down{(-45.45)} & 35.23 \down{(-59.82)} \\   \bottomrule
\end{tabular}
\end{subtable}
\begin{subtable}{1\linewidth}
\centering
\begin{tabular}{ccccp{23.5mm}<{\centering}p{23.5mm}<{\centering}p{23.5mm}<{\centering}}
\toprule
\multirow{2}{*}{\textbf{Dataset}} & \multirow{2}{*}{\textbf{Method}} & \multirow{2}{*}{\textbf{ Scheme}} & \multirow{2}{*}{\textbf{Acc. (Authorized)}} & \multicolumn{3}{c}{\textbf{Acc. (Hacker)}}  \\ \cmidrule{5-7} 
                           &                         &                                  &                                  & MobileNetV2 & ShuffleNetV2  & ResNet-18 \\ \midrule\midrule
CIFAR-100 & Original & Normal & 78.24  & 68.92 & 71.26 & 78.24  \\
CIFAR-100  & Original & Distill & - & 72.67 \down{(+3.75)} & 74.39 \down{(+3.13)} & 79.24 \down{(+1.00)}  \\ \midrule
CIFAR-100  & UGEs w/o UD  & Normal  & 75.26 \down{(-2.98)}& 22.05 \down{(-46.87)}  &  21.59 \down{(-49.67)}  & 16.46 \down{(-61.78)}   \\
CIFAR-100  & UGEs w/o UD  & Distill &  -& 63.52 \down{(-5.40)} & 58.23 \down{(-13.03)} &40.45 \down{(-37.79)}\\ \midrule
CIFAR-100  & UGEs  & Normal     &  74.68 \down{(-3.56)}& 32.11\down{(-36.81)}& 28.33\down{(-42.93)} &16.55 \down{(-61.69)} \\
CIFAR-100  & UGEs  & Distill    &  -  &26.94 \down{(-41.98)}&25.34\down{(-45.92)} &15.50 \down{(-62.74)} \\ \midrule \midrule
TinyImageNet  & Original   & Normal  &  63.08  &  56.00& 59.90& 63.08       \\ 
TinyImageNet  & Original    & Distill   & - & 60.02 \down{(+3.06)} & 63.19 \down{(+7.19)}  &66.28 \down{(+6.38)}  \\ \midrule
TinyImageNet  & UGEs w/o UD & Normal  & 60.88 \down{(-2.20)}&14.62 \down{(-41.38)} &13.97 \down{(-45.93)} &15.34 \down{(-47.74)}     \\
TinyImageNet   & UGEs w/o UD  & Distill  & - & 36.39 \down{(-6.18)} &49.82 \down{(-16.97,)} &42.93 \down{(-20.15)}  \\ \midrule
TinyImageNet    & UGEs   & Normal   & 59.54 \down{(-3.54)}&15.79 \down{(-40.21)} &22.75 \down{(-37.15)}&17.55 \down{(-45.53)}      \\
TinyImageNet   & UGEs  & Distill   & -     & 19.69 \down{(-36.94)} &21.06 \down{(-35.48)} &24.42 \down{(-38.66)}      \\
\bottomrule
\end{tabular}
\end{subtable}
\label{tab:abl}
\end{table*}

\subsection{Experimental Results}
\textbf{Ablation Study.}
The results of the ablation study on CIFAR-10, CIFAR-100 and TinyImageNet datasets are presented in Table~\ref{tab:abl}. We compare the test accuracy on both the authorized network (Acc. (Authorized)) and the hacker networks (Acc. (Hacker)). Various network backbones are chosen to create the hacker networks, trained under two schemes: normal training with data labels (Normal) and distillation-based training using Eq.~\ref{eq:distill} (Distill).
The distillation-based training can be thought as a kind of attack.
For comparison, we show and compare the results with:
`Original': the unmodified original data $\mathcal{D}$;
`Unlearn': training only with unlearn loss $\mathcal{L}_{fd}$;
`UnDistill': training only with the undistillaion loss $\mathcal{L}_{ud}$;
`UGEs w/o UD': training without the undistillaion loss $\mathcal{L}_{ud}$.
Form the table, conclusions could be drawn that:
\begin{itemize}
    \item The efficacy of UGEs is evaluated based on their learnability on the authorized network (higher Acc. (Authorized)) and unlearnability on hacker networks (lower Acc. (Hacker)). Our method significantly reduces the test accuracy of hacker networks (by more than 40\%) while maintaining an acceptable drop in authorized network accuracy (less than 5 \%).
    \item UGE effectiveness is demonstrated across CIFAR-10, CIFAR-100, and TinyImageNet datasets, utilizing diverse network architectures like ResNet-18, CNN, MobileNet, and ShuffleNet. This showcases UGEs' versatility and efficacy across different scenarios;
    \item Our proposed UGEs demonstrate robustness against attacks where hackers use the authorized network to acquire the learnability of UGEs (Scheme as `Distill'). The results show that the proposed undistillation loss $\mathcal{L}_{ud}$ (comparing `UGEs' and `UGEs w/o UD' ) effectively prevents such attacks, fulfilling the robustness goals in the design.
\end{itemize}

\begin{table}[t]
\small
\caption{Results on UGEs with multiple authorized networks on CIFAR-10 dataset, which are tested under three training schemes.}
\begin{tabular}{cccccc}
\toprule
\multirow{2}{*}{\textbf{Method}} & \multirow{2}{*}{\textbf{Scheme}} & \multicolumn{2}{c}{\textbf{Authorized}} & \multicolumn{2}{c}{\textbf{Hacker}} \\ \cmidrule(r){3-4} \cmidrule(r){5-6} 
                        &                         & Net-1             & Net-2             & CNN          & ResNetC-20         \\ \midrule\midrule
Original  & Normal  &  95.01    &  94.95    &  86.57  & 93.04        \\
Original   & Distill-1    &   -    & -    &     88.06    &   92.62     \\
Original   & Distill-2    &  -  &  - &   88.09   &   92.55        \\ \midrule
UGEs      & Normal     &  93.27   &  93.83  &   43.28  &  49.34                \\
UGEs     & Distill-1   & -    &   -   &  53.60  &  56.32                  \\
UGEs    & Distill-2   &   -   &   -   &   50.32  &  54.29           \\ \bottomrule
\end{tabular}
\label{tab:multi}
\end{table}

\textbf{UGEs with Multiple Authorized Networks.}
In the standard experimental setup, we initially configure one network as the authorized network. Here, we extend our framework to accommodate multiple authorized networks, introducing additional loss items for each newly added network in $\mathcal{L}_{all}$. Refer to the supplementary material for specific details on modifying the losses and extra experiments.

In this extension, we establish two authorized networks with ResNet-18 with distinct initialization parameters. The experimental results on the CIFAR-10 dataset are presented in Table~\ref{tab:multi}, where `Distill-1' indicates optimizing the hacker networks with the distillation calculated on authorized net-1.
As observed in the table, introducing another authorized network maintains the effectiveness of our proposed framework for learning UGEs. However, with more authorized networks, the performance of UGEs experiences a slight decline. Addressing this challenge and providing a more flexible framework to include multiple authorized networks will be a focus of our future work.

\begin{table}[t]
\centering
\small
\caption{Comparing the data unlearnability with the existing ULE methods on CIFAR-10 and CIFAR-100 datasets.}
\begin{tabular}{p{18mm}<{\centering}p{11mm}<{\centering}p{11mm}<{\centering}p{11mm}<{\centering}p{11mm}<{\centering}}
\toprule
\multirow{2}{*}{\textbf{Method}} & \multicolumn{2}{c}{\textbf{Acc. (CIFAR-10)}} & \multicolumn{2}{c}{\textbf{Acc. (CIFAR-100)}} \\ \cmidrule(r){2-3}  \cmidrule(r){4-5} 
                        & \textbf{Clean}  & \textbf{ULEs}  & \textbf{Clean} & \textbf{ULEs}    \\ \midrule\midrule
EM~\cite{huang2020unlearnable}     & 94.66         & 13.20        & 76.27         & 1.60          \\
TAP~\cite{fowl2021adversarial}    & 94.66         & 22.51        & 76.27         & 13.75         \\
NTGA~\cite{yuan2021neural}   & 94.66         & 16.27        & 76.27         & 3.22          \\
REM~\cite{fu2021robust}                    & 94.66         & 27.09        & 76.27         & 10.14         \\
CUDA~\cite{sadasivan2023cuda}      & 94.66         & 18.48        & 76.27         & 12.69     \\
Ours ($\mathcal{L}_{fd}$) & 94.66& 22.59 & 76.27 & 9.35 \\ \bottomrule
\end{tabular}
\label{tab:unlearn}
\end{table}
\begin{figure}[t]
\centering
\includegraphics[scale = 0.6]{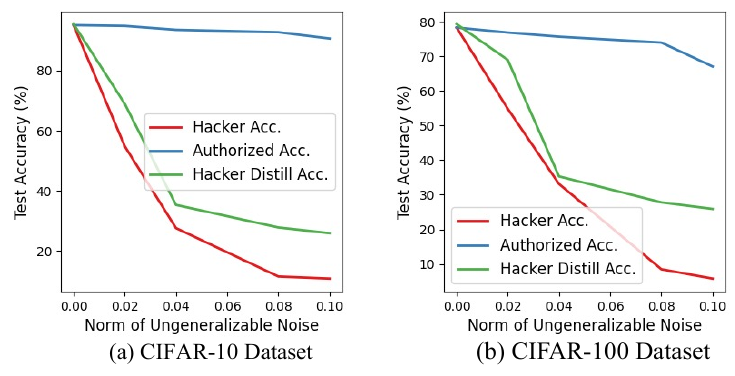}
\caption{
The performance concerning the value of $\rho$ on CIFAR-10 and CIFAR-100 datasets. }
\label{fig:norm}
\end{figure}

\textbf{How does the norm of ungeneralizable noise affect the UGEs performance.}
Recall that we set the norm of the ungeneralizable noise $\rho$ as 0.04. We investigate the performance concerning the value of $\rho$, as illustrated in Fig.~\ref{fig:norm}.
From the figure, it can be observed that a larger norm of ungeneralizable noise leads to a decrease in test accuracy on the authorized network. Therefore, a properly chosen small norm of noise is essential, ensuring both the visual integrity of the protected data and maintaining acceptable authorized network performance.

\textbf{Comparing with Existing ULEs.}
We compared the proposed method with existing ULE methods on CIFAR-10 and CIFAR-100 datasets, as shown in Table~\ref{tab:unlearn}. The listed methods are included in the table, and the experimental setup follows previous work~\cite{sadasivan2023cuda}. Lower test accuracy indicates better unlearnability. The results demonstrate that our proposed method contributes to current unlearnable example methods, achieving competitive results with existing ULE methods. The UGE framework can also be seamlessly integrated into the ULEs framework by training the generator $\mathcal{G}$ with the loss term $\mathcal{L}_{df}$.

\begin{figure}[t]
\centering
\includegraphics[scale = 0.58]{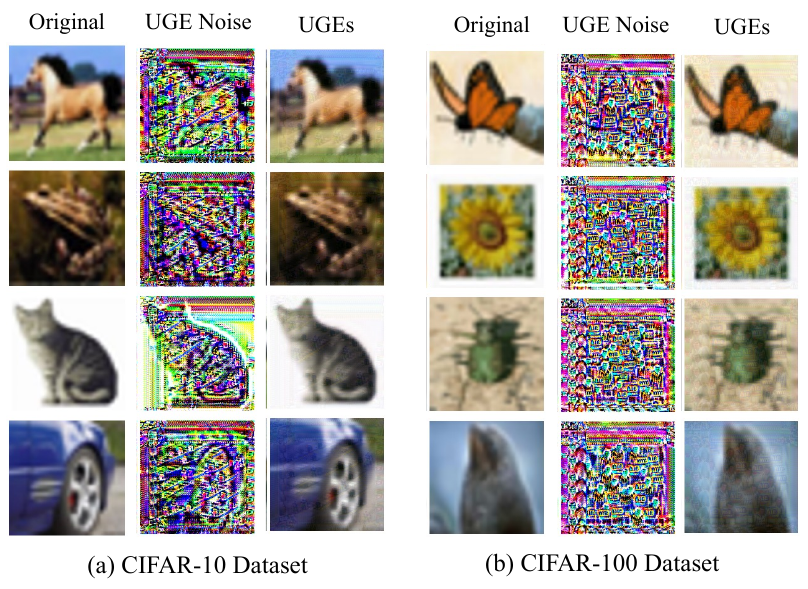}
\caption{
The visualization results include the original clean images, the ungeneralizable noise (scaled by $255$ for better visualization), and the resultant ungeneralizable images.
}
\label{fig:vis}
\end{figure}

\textbf{More Analysis.}
In Fig.~\ref{fig:vis}, we showcase the visualization results of our proposed UGE. The UGEs demonstrate visual similarity to the original images, confirming their visual integrity and aligning with the framework's design goal.

\subsection{Limitations}
While our method shows promise across various scenarios, it has limitations, particularly when faced with an increasing number of authorized networks (Table.~\ref{tab:multi}). Addressing this, we plan to incorporate ensemble methods or knowledge amalgamation to enhance UGEs' performance in such scenarios. This underscores our commitment to ongoing improvement and adaptability.
It's important to note that our UGE framework is designed for classification tasks. Looking ahead, we aim to extend its applicability to multiple tasks, enabling the seamless transition of data learnability among different tasks, thus enhancing its versatility.

%% file: sec/5_conclusion.tex
\section{Conlusion}
In conclusion, our paper presents the ungeneralizable examples framework, a versatile paradigm for data protection. UGE allows legitimate data usage by the protector while preventing unauthorized access by potential hackers. The proposed approach, incorporating three distinct losses, successfully achieves a seamless transition between data learnability and unlearnability. Empirical verification validates the effectiveness and robustness of our method, demonstrating its potential in enhancing data security in machine learning applications.


%% file: sec/acknowledge.tex
\section*{Acknowledgements}
This project is supported by 
the Advanced Research and Technology Innovation Centre (ARTIC), the National University of Singapore under Grant (project number: A0005947-21-00, project reference: ECT-RP2).

%% file: sec/X_suppl.tex
\clearpage
\setcounter{page}{1}
\maketitlesupplementary
In this document, we present supplementary materials that couldn't be accommodated within the main manuscript due to page limitations. Specifically, we offer additional details on the proposed UGE framework, including the architecture of the generator and the modified losses with multiple authorized networks and  concrete experimental setting.

\section{More Details of UGEs}

To optimize the UGEs, we utilize a total loss consisting of three distinct components. Below, we provide more details on constructing the UGE framework for multiple authorized networks, as well as specifics regarding the generator for the ungeneralizable noise.

\subsection{UGEs with Multiple Authorized Networks}
It is mentioned in the main paper that we consider the scenario with one authorized network. However, we assert that our proposed framework is capable of handling cases where multiple authorized networks are determined by the protector.

Denote the authorized network set as: $F = \{f^1_\theta,f^2_\theta,...,f^K_\theta \}$, then for each loss item in $\mathcal{L}_{all}$, each could be rewritten as:
\begin{equation}
\begin{split}
  \mathcal{L}_{gm} =\frac{1}{|\tau|\!\times\! n\times K}\sum_{f^k\in F}\sum_{t\in \tau} &\sum_{(x,y)\in \mathcal{D}}\mathcal{D}{ist}\Big[\nabla\mathcal{L}\big(f^k_{\theta_t}(x),y\big),\\
  &\nabla\mathcal{L}\big(f^k_{\theta_t}(x_u),y\big)\Big],  \\
\mathcal{L}_{ud}\! =\! \frac{1}{n\times K}\!\sum_{f^k\in F}\sum_{(x_u,y_u)\in \mathcal{D}_u}\!&\big[ \mathcal{L}(f^k_\theta(x_u), y_u)
    \\
    -\!  \omega \mathcal{L}_{kd} \big(&f^k_\theta(x_u), f'_{\theta_A}(x_u)\big)\big].
\end{split}
\end{equation}
Here, we modify the loss items of $\mathcal{L}_{gm}$ and $\mathcal{L}_{ud}$ while keeping the loss item $\mathcal{L}_{fd}$ unchanged.
It is important to note that as the total number of authorized networks increases, the performance of our synthetic method may decrease. Detailed experiments regarding the number $K$ are conducted in the subsequent section.

\subsection{The Architecture of the Generator}
\begin{figure}[t]
\centering
\includegraphics[scale = 0.45]{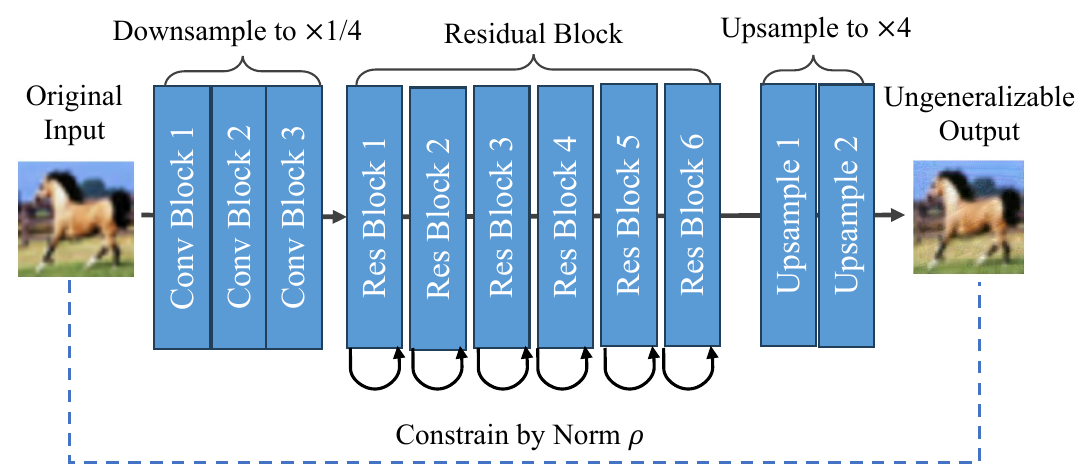}
\caption{
The architecture of the generator to synthesize the ungeneralizable examples. }
\label{fig:gan}
\end{figure}

Note that in the main paper, we utilize the generator $\mathcal{G}$ to synthesize the ungeneralizable version of the data, which is denoted as:
\begin{equation}
    x_u = \mathcal{G}(x), \quad x\in \mathcal{D},
\end{equation}
where we omit the operation to constrain the norm of the ungeneralizable noise. And the ungeneralizable examples $x_u$ form the final published ungeneralizable dataset.
We use the ResNet based backbone for constructing the generator. To be concrete, the architecture of the generator is given in Table.~\ref{tab:generator} and Fig.~\ref{fig:gan}, which consists of conv, residual and upsampling blocks.

\begin{table}[t]
\centering
\small
\caption{The architecture of the generative perturbation network.}
\begin{tabular}{ccc}
\toprule
\textbf{Block Name} & \textbf{Output Size}& \textbf{Layers} \\\midrule \midrule
Conv Block & $56\times 56 $ & $\begin{bmatrix} \text{Conv Layer ($7\times 7$)}\\\text{BatchNorm} \\ \text{LeakyReLU}\end{bmatrix}\times 1$ \\ \midrule
Conv Block & $56\times 56 $ & $\begin{bmatrix} \text{Conv Layer ($3\times 3$)}\\\text{BatchNorm} \\ \text{LeakyReLU}\end{bmatrix}\times 2$ \\ \midrule
Res Block  & $56\times 56 $  &  $\begin{bmatrix} \text{Conv Layer ($3\times 3$)}\\\text{BatchNorm} \\ \text{LeakyReLU}\\ \text{Dropout}\\ \text{Conv Layer ($3\times 3$)} \\ \text{BatchNorm}\end{bmatrix}\times 6$\\\midrule
Upsample   & $224\times 224 $  & 
    $\begin{bmatrix} \text{Conv Transpose($3\times 3$)}\\\text{BatchNorm} \\ \text{LeakyReLU}\end{bmatrix}\times 2$ \\
    &&   \quad Conv Layer\quad \quad\quad \\
\bottomrule
\end{tabular}
\label{tab:generator}
\end{table}

\begin{figure*}[t]
\centering
\includegraphics[scale = 0.5]{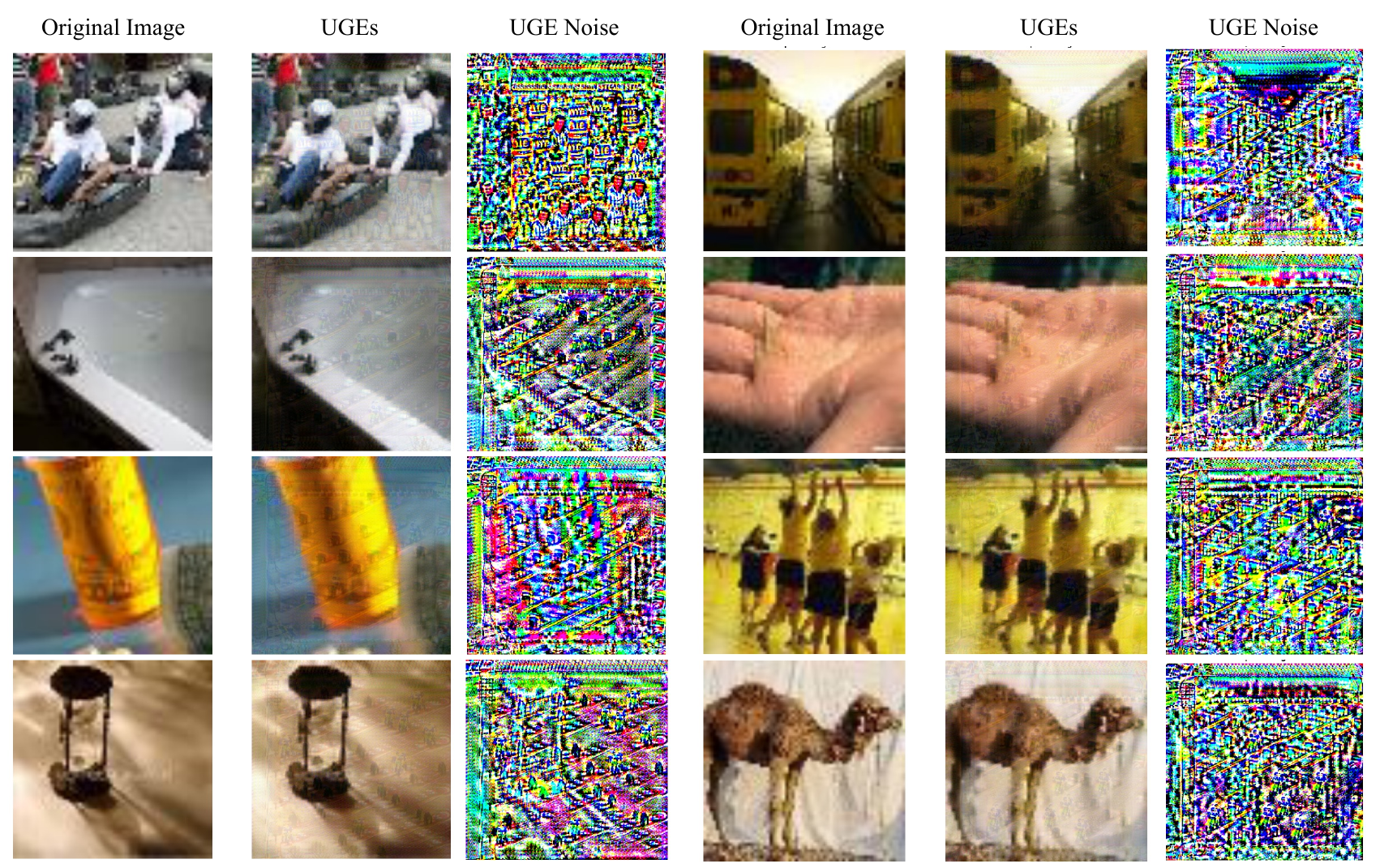}
\caption{
 The visualization results include the original clean images, the ungeneralizable noise (scaled by 255 for better visualization), and the resultant ungeneralizable images. These visualizations are presented for TinyImageNet dataset.}
\label{fig:visimage}
\end{figure*}

\begin{table}[t]
\centering
\small
\caption{The hyper parameters setting for CIFAR-10, CIFAR-100 and TinyImageNet datasets.}
\begin{tabular}{cccccc}
\toprule
\multirow{2}{*}{\textbf{Dataset}} & \multicolumn{2}{c}{$\mathcal{L}_{all}$} & $\ell_{tri}$ & \multicolumn{2}{c}{$\mathcal{L}_{ud}$} \\\cmidrule(r){2-3}\cmidrule(r){4-4}\cmidrule(r){5-6}
                         & $\lambda_{fd}$          & $\lambda_{ud}$  & $\alpha$  & $\omega$            & $T$          \\\midrule \midrule 
CIFAR-10                 & 1           & 0.1          & 0.1    & 0.1          & 4          \\
CIFAAR-100               & 1           & 0.1          & 0.1    & 0.1          & 20         \\
TinyImageNet             & 1           & 1            & 0.1    & 0.01         & 20        \\\bottomrule
\end{tabular}
\label{tab:params}
\end{table}

\begin{table}[t]
\centering
\small
\caption{The training details for generator and normal networks on CIFAR-10, CIFAR-100 and TinyImageNet datasets.}
\begin{tabular}{cccc}
\toprule
\textbf{Dataset}                       & \textbf{Network}        & \textbf{Learning Rate} & \textbf{Epoch} \\ \midrule\midrule
\multirow{3}{*}{CIFAR-10}     & Generator      & 1e-3          & 50    \\
                              & Plain CNN      & 1e-3          & 100   \\
                              & Other Networks & 0.1           & 160   \\\midrule
\multirow{2}{*}{CIFAR-100}    & Generator      & 1e-3          & 100   \\
                              & Networks       & 0.1           & 200   \\\midrule
\multirow{2}{*}{TinyImageNet} & Generator      & 1e-3          & 100   \\
                              & Networks       & 0.1           & 200  \\\bottomrule
\end{tabular}
\label{tab:net}
\end{table}

\section{Experiments Setup}
Here is a detailed setting for each part of the experiments.

The balancing weights and other hyperparameters in each loss item are provided in Table~\ref{tab:params}, specifying the parameter settings for CIFAR-10, CIFAR-100, and TinyImageNet datasets, respectively.

And the details regarding the network training are given in Table~\ref{tab:net}, where the training of the generator and the normal networks is given.


\section{More Experimental Results}

\subsection{More Visualization Results on TinyImageNet}
In the main paper, we presented visualizations of ungeneralizable examples on CIFAR-10 and CIFAR-100 datasets. Here, we provide additional visualization results on the TinyImageNet dataset, as shown in Fig.~\ref{fig:visimage}. 
We also visualize the ungeneralizable noise, which could reflect some details of the original image, showing that the learned UGE noise is sample-wise.
The figure illustrates that our proposed UGE framework is capable of generating visually integrated ungeneralizable images from the original inputs, demonstrating its effectiveness on more complex datasets.

\subsection{UGEs with Multiple Authorized Networks}
We already give the experimental results on UGEs with multiple authorized networks on CIFAR-10 dataset in Table~\ref{tab:multi}.
In this experiments, we set two authorized networks with the same architecture (ResNet-18) but with different kinds of initialization.
Here we explore deeper on the mutiple authorized networks cases.

\textbf{Effect of the Number of Authorized Networks on UGEs Performance}
In this experiment, we investigate the impact of the number of authorized networks on the performance of UGEs. Specifically, we conduct the experiment using three authorized networks, all sharing the same architecture (ResNet-18) but initialized with different parameters.
The experimental results are depicted in Table~\ref{tab:multi}, where we can observe that:
\begin{itemize}
    \item The effectiveness of the proposed UGEs is further demonstrated in a scenario involving three authorized networks (`Net-1', `Net-2' and `Net-3'). In this case, the UGEs achieve approximately $90\%$ test accuracy on the authorized networks and around $70\%$ test accuracy on the hacker network  (`CNN').
    \item Nevertheless, it's important to note a slight decrease in test accuracy on the authorized networks as the number of authorized networks increases. Specifically, the test accuracy is observed to be $93.89\%$ for one authorized network, $93.55\%$ for two authorized networks, and $90.09\%$ for three authorized networks. Concurrently, the test accuracies on the hacker network show an increase with the addition of more authorized networks.
\end{itemize}

\begin{table}[t]
\small
\caption{Results on UGEs with multiple authorized networks on CIFAR-10 dataset, which are tested under three training schemes.}
\begin{tabular}{cccccc}
\toprule
\multirow{2}{*}{\textbf{Method}} & \multirow{2}{*}{\textbf{Scheme}} & \multicolumn{3}{c}{\textbf{Authorized}} & \multicolumn{1}{c}{\textbf{Hacker}} \\ \cmidrule(r){3-5} \cmidrule(r){6-6} 
                        &                         & Net-1             & Net-2      & Net-3          & CNN           \\ \midrule\midrule
Original  & Normal  &  95.01    &  95.01  & 95.06 &  86.57        \\
Original   & Distill-1    &   -    & -   &- &     88.06       \\
Original   & Distill-2    &  -  &  - & - & 88.09        \\
original & Distill-2    &  -  &  - & - & 88.14  \\  \midrule
UGEs      & Normal     &  90.39   &  89.64  &90.24 &  72.42                \\
UGEs     & Distill-1   & -    &   -   &-  & 73.60                 \\
UGEs    & Distill-2   &   -   &   -   &-  & 75.32           \\ 
UGEs    & Distill-3   &   -   &   -   &-  &  74.32          \\ \bottomrule
\end{tabular}
\label{tab:three}
\end{table}

\begin{figure}[t]
\centering
\includegraphics[scale = 0.45]{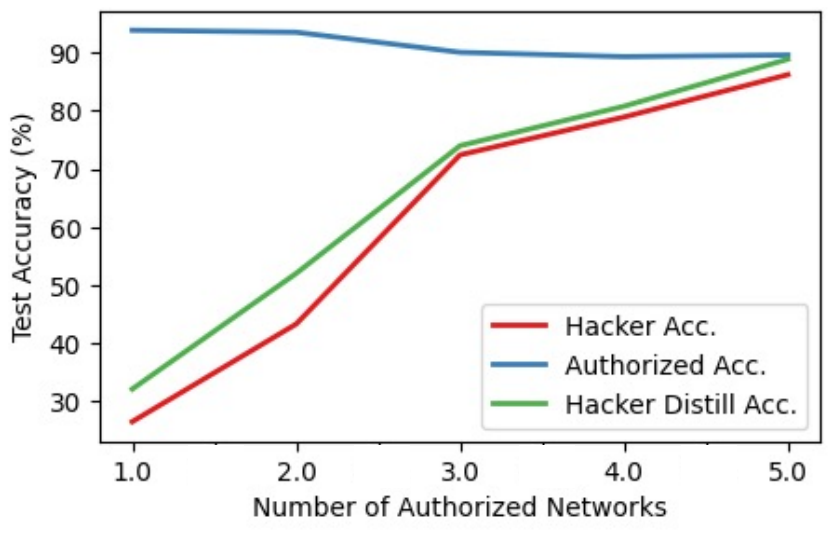}
\caption{The performance of the proposed UGEs regarding the total number of the authorized networks. The `Authorized Acc.' is calculated on the average test accuracy on all the authorized networks, which is similar for `Hacker Distill Acc.'.}
\label{fig:number}
\end{figure}

Additionally, we analyze the relationship between the number of authorized networks and the corresponding test accuracies. To achieve this, we calculate the average test accuracies across multiple authorized networks, while employing the test accuracy obtained from the plain CNN as the representative hacker network. The results of this analysis are illustrated in Fig.~\ref{fig:number}.

\begin{table}[t]
\small
\caption{Results on UGEs with multiple authorized networks on CIFAR-10 dataset, which are tested under three training schemes.}
\begin{tabular}{cccccc}
\toprule
\multirow{2}{*}{\textbf{Method}} & \multirow{2}{*}{\textbf{Scheme}} & \multicolumn{2}{c}{\textbf{Authorized}} & \multicolumn{2}{c}{\textbf{Hacker}} \\ \cmidrule(r){3-4} \cmidrule(r){5-6} 
                        &                         & CNN            & Res-18             & CNN          & Res-18        \\ \midrule\midrule
Original  & Normal  &  86.24    &  95.01    &  86.57  & 95.06       \\
Original   & Distill-1    &   -    & -    &     87.51    & 95.10     \\
Original   & Distill-2    &  -  &  - &   88.06  &   95.44        \\ \midrule
UGEs      & Normal     &  82.95  &  93.08  &   45.26  &  50.43               \\
UGEs     & Distill-1   & -    &   -   &  48.92  &  55.06                  \\
UGEs    & Distill-2   &   -   &   -   &   48.57  &  54.74           \\ \bottomrule
\end{tabular}
\label{tab:diffmulti}
\end{table}

\textbf{Performance of UGEs on Multiple Authorized Networks with Different Architectures}
We further investigate the applicability of UGEs in scenarios with multiple authorized networks employing different architectures. In this experiment, we choose the plain CNN and ResNet-18 to constitute the set of authorized networks. The experiments are conducted on the CIFAR-10 dataset, and the results are detailed in Table~\ref{tab:diffmulti}.

From the table, we observe that:
\begin{itemize}
    \item The effectiveness of our proposed UGEs extends to scenarios with multiple authorized networks employing different architectures. In this experiment, utilizing both plain CNN and ResNet-18 as authorized networks on the CIFAR-10 dataset, we observe that the test accuracy on authorized networks drops by less than $4\%$. Conversely, the test accuracies for hacker networks experience a significant reduction of more than $40\%$.
    \item In comparison to scenarios with multiple authorized networks sharing the same architecture, the UGEs with different architectures for authorized networks show a slight drop in performance. This discrepancy is primarily attributed to the strict trajectory alignment. Addressing this challenge presents a potential avenue for future improvements to enhance the UGE framework.
\end{itemize}

\begin{table*}[t]
\small
\centering
\caption{Applying UGEs in federated learning setting. The experiments are conducted on CIFAR-10 dataset. We use the plain CNN and ResNet-18 as the hacker networks.}
\begin{tabular}{cccccccccccc}
\toprule
\multirow{2}{*}{\textbf{Acc.}} & \multicolumn{2}{c}{\textbf{Normal Training}} & \multicolumn{3}{c}{\textbf{Federated}} & \multicolumn{6}{c}{\textbf{Hackers}}    \\ \cmidrule(r){2-3}\cmidrule(r){4-6} \cmidrule(r){7-12}
 & Joint         & Separate          & Sever1   & Sever2   & Global  & CNN-N & CNN-H & CNN-D &  Res18-N & Res18-H & Res18-D \\ \midrule\midrule
Acc. F5    & 93.83   & 93.54      & 92.91    & -     & 91.37   & 85.18   & 48.46 & 50.46   & 93.77      & 69.17     & 71.19             \\
Acc. L5   & 96.27   & 96.50   & -   & 96.18    & 95.50   & 86.05      & 49.70 & 58.45       & 96.31    & 70.02     & 73.02             \\\midrule
Avg. Acc.  & 95.05   & 95.02   & -  & -        & 93.44   & 85.62      & 49.08 & 54.46       & 95.04    & 69.60     & 72.11    \\ \bottomrule
\end{tabular}
\label{tab:fl}
\end{table*}

\subsection{UGEs for Federated Learning}
In Sec.~\ref{sec:use}, we assert the practicality and deployability of the proposed UGE framework across various applications. To illustrate, consider a scenario where a global server establishes the global network as $f_\theta$, and two local servers, each possessing its distinct dataset—$\mathcal{D}^1$ for the first server and $\mathcal{D}^2$ for the second server. The global network is supposed to train on the two datasets $\mathcal{D}^1 \cup \mathcal{D}^2$, while not requiring the data shared between each server.

In this setup, both servers can independently generate their versions of UGEs with the information of the global network $f_\theta$, denoted as:
\begin{equation}
\begin{split}
    \mathcal{D}_u^1 \leftarrow \arg\min_{x_u} \mathcal{L}_{all}(\mathcal{D}^1, f_\theta),\\
    \mathcal{D}_u^2 \leftarrow \arg\min_{x_u} \mathcal{L}_{all}(\mathcal{D}^2, f_\theta),
\end{split}  
\end{equation}
where the generation of $\mathcal{D}_u^1$ and $\mathcal{D}_u^2$ involves no data interaction, ensuring data privacy within each local server, meeting the basic privacy concern of federated learning.

After each server uploads its ungeneralizable version of the data, the global model can be jointly trained as follows:
\begin{equation}
    f: \min_\theta \frac{1}{|\mathcal{D}_u^1|+ |\mathcal{D}_u^2|}\sum_{\{x_u,y_u\}\in \mathcal{D}_u^1\cup \mathcal{D}_u^2 }\mathcal{L}(f_\theta(x_u),y_u).
\end{equation}
This optimization involves training the network using a normal training scheme with the combined datasets from both servers.

In order to test the effectiveness of UGEs applied in federated learning, we designed the experiment as follows. We selected ResNet-18 as the global model and divided the CIFAR-10 dataset into two parts. The first part includes data for the first 5 classes and is hosted by local server 1, while the second part includes data for the remaining 5 classes and is hosted by local server 2.

The experimental results are depicted in Table~\ref{tab:fl}, where we compare the accuracies of the first 5 classes (`Acc. F5'), accuracies of the last 5 classes (`Acc. L5') and the average accuracy across all 10 classes (`Avg. Acc.').
pecifically, the methods for comparison include: (1) networks with normal training, trained on the total dataset $\mathcal{D}^1\cup \mathcal{D}^2$ (`Joint') and trained on each sub-dataset separately (`Separate'); (2) networks in a federated learning setting, including the authorized network trained on $\mathcal{D}_u^1$ (`Server1'), the authorized network trained on $\mathcal{D}_u^2$ (`Server2'), and the authorized network trained on $\mathcal{D}_u^1\cup\mathcal{D}_u^2$ (`Global'); (3) hacker networks with a CNN backbone trained with $\mathcal{D}^1\cup\mathcal{D}^2$ (`CNN-N'), trained on $\mathcal{D}_u^1\cup\mathcal{D}_u^2$ (`CNN-H') and that distilled from the authorized network (`CNN-D'), with a ResNet-18 backbone trained with $\mathcal{D}^1\cup\mathcal{D}^2$ (`Res18-N'), trained with $\mathcal{D}_u^1\cup\mathcal{D}_u^2$ (`Res18-H') and that distilled from the authorized network (`Res18-D').

From the table, we observe that:
\begin{itemize}
    \item Our proposed UGE framework precisely aligns with the privacy requirements of federated learning, preventing shared data from being reused by third parties and prohibiting data interaction between each local server.
    \item The generated UGEs not only work when locally training the network $f_\theta$ ( `Sever1'\& `Sever2' with less than $1\%$ accuracy drop), but also when jointly training the global network $f_\theta$ ( `Global' with less than $2\%$ accuracy drop).
    \item The generated UGEs effectively prevent reuse by hacker networks, resulting in reduced accuracies on CNN from $85.62\%$ to $49.08\%$ and on ResNet-18 from $95.04\%$ to $69.60\%$. Additionally, it mitigates information leakage from the global network, as the network after distillation shows a relatively low accuracy compared to normal training (a $20\%$ to $30\%$ accuracy drop).
\end{itemize}